\documentclass{article}
\usepackage{arxiv}

\usepackage[round]{natbib}

\usepackage{microtype}
\usepackage{graphicx}
\usepackage{subfigure}
\usepackage{booktabs} 

\usepackage{hyperref}

\usepackage{amsmath}
\usepackage{amssymb}
\usepackage{mathtools}
\usepackage{amsthm}

\usepackage[capitalize,noabbrev]{cleveref}

\theoremstyle{plain}

\theoremstyle{definition}

\theoremstyle{remark}

\usepackage[textsize=tiny]{todonotes}

\title{Generation Constraint Scaling Can Mitigate Hallucination}

\date{}
\author{
  Georgios Kollias, 
  Payel Das, 
  Subhajit Chaudhury \\ 
  IBM Research\\
  T. J. Watson Research Center\\
  \texttt{\{gkollias, daspa\}@us.ibm.com, subhajit@ibm.com}
}

\renewcommand{\headeright}{}
\renewcommand{\undertitle}{}

\begin{document}
\maketitle

\begin{abstract}
Addressing the issue of hallucinations in large language models (LLMs) is a critical challenge. As the cognitive mechanisms of hallucination have been related to memory, here we explore hallucination for LLM that is enabled with   explicit memory  mechanisms.
We empirically demonstrate that by simply scaling the readout vector that constrains generation in a memory-augmented LLM decoder,  hallucination mitigation can be achieved in a training-free manner. 
Our method is geometry-inspired and outperforms a state-of-the-art LLM editing method on the task of generation
of Wikipedia-like biography entries both in terms of generation quality and runtime complexity.   
\end{abstract}

\section{Introduction and Background}
While large language models exhibit remarkable performance in language generation and machine translation, their generations suffer from the issue of hallucinations. Model editing techniques provide a path to mitigate such issues, which involve modifying model parameters such that  model outputs are changed to desired responses for specific questions without compromising the accuracy for others. Context-grounding has been proposed as another means, where desired response (the actual fact) is included in the context within the prompt, with the expectation that the decoder will utilize the information included in the prompt. Given the relation between memory and hallucination in psychology \cite{berberette2024redefining}, it is believed that LLMs with explicit memory mechanism will help lowering hallucination. Here, we investigate if that is indeed the case, by employing Larimar \cite{das2024larimar}, a recently proposed  LLM decoder that is augmented with an external memory with read/write access.
In the memory-augmented LLM, the encodings of arguments to their memory primitives serve only as intermediate representations in their generation pipeline 
to condition the decoding: and then
they are silently discarded, they are not further explored. Departing from this practice, we inspect the geometry of these representations and leverage our findings to devise a simple yet effective approach for mitigating hallucination.  Here we use Larimar as the memory-augmented language model and compare its performance on a hallucination benchmark with GRACE \cite{hartvigsen2022aging} as a baseline, which is an existing model editing technique.

\subsection{Larimar}

Larimar 
is a class of LLMs augmented with an external episodic memory controller. In its base instantiation, Larimar architecture includes (i) an encoder, (ii) an associative memory module and (iii) a decoder. The encoder computes latent representations of sets of textual inputs (episodes) and queries. These can be respectively used for updating the memory and querying it to return readout encodings. The decoder generates output text from a prompt, constrained by the readout.

Note that a readout vector serves as a special compressed key-value (KV) cache at the decoder, which expands to key-value vector pairs for each of its layers via a weight matrix learnt during Larimar training. It can also be interpreted as the latent vector that is injected for adapting the decoder to arbitrary conditional input without retraining the model again as in Optimus \cite{li2020optimus}.

\subsection{GRACE}
GRACE 
is a method for LLM editing without altering its weights. It works by installing a special adapter at one or more of its layers: a GRACE adapter is basically a dynamically expanding key-value codebook. A key is the layer's input activation; its value is the input to the next layer that, if substituted, would render the correct output for the current input-output sample pair. Codebook values are learnt by minimizing a task-specific loss function.

\section{Experiments}

\textbf{Data and Models}

WikiBio is a 
hallucination benchmark
dataset of Wikipedia-like biographies for 238 subjects generated by prompting GPT-3 \cite{manakul2023selfcheckgpt}. It includes annotations for the factuality accuracy for each of the generated sentences by comparing them to the actual Wikipedia biography article sentences (accurate, major/minor inaccurate) \footnote{\url{https://huggingface.co/datasets/potsawee/wiki_bio_gpt3_hallucination}}. 

In \cite{hartvigsen2022aging}, authors finetune GPT2-XL on WikiBio dataset mixed with sentences from OpenWebText \cite{openwebtext}, a public version of GPT2’s training data and use the resulting model
\footnote{\url{https://huggingface.co/tomh/gpt2xl-grace}} 
in hallucination mitigation experiments. We apply edits on the same model and adapter configuration for GRACE benchmarks. 

For Larimar-based experiments, we employ Larimar-1.3B model. This comprises a BERT large encoder~\cite{devlin2018bert} combined with a GPT2-large~\cite{radford2019language} decoder and a memory matrix (512x768), trained over 7.6 million examples constructed by splitting WikiText ~\cite{merity2016pointer} texts to small chunks of 64 tokens.

\textbf{WikiBio hallucination task}

We organize the sequence of $n_i$ sentences in the actual and GPT-3 generated (\texttt{hal}lucinating) texts for the $i^{th}$ WikiBio entry, $i\in [238]$, in two lists: $[\mathbf{WB}_i(j)]$ and $[\mathbf{WB}_i^{\texttt{hal}}(j)]$, $j\in [n_i]$. Then for each WikiBio entry $i$ we form $n_i - 1$ sentence pairs  of the form $(\mathbf{WB}_i^{\texttt{hal}}(j), \mathbf{WB}_i(j + 1)))$, $j\in[n_i-1]$. The first sentence in the pair (also referred to as \texttt{prompt} in the sequel) is a GPT-3 generated sentence (with index $j$) in the entry  and the second sentence (alternate name: \texttt{input}) is the ``next'' one (index: $j+1$), however the latter in the sequence of the sentences in the \emph{actual} Wikipedia entry.

We can then use these pairs to generate output sentences from a model. In particular, we:

\begin{enumerate}
\item
\textbf{Inform} the model of a pair (\texttt{prompt}, \texttt{input}). In Larimar this can be achieved by writing this pair to its memory. For GRACE this will be an edit operation typically updating the codebook in one of its adapters.   
\item
\textbf{Ask} the informed model to generate an \texttt{output} sentence based on the \texttt{prompt} only; \texttt{input} is no longer accessible. In Larimar this will be implemented by (i) querying the memory with the prompt to get a \texttt{readout} vector and (ii) generating model \texttt{output} initialized to \texttt{prompt} and constrained by \texttt{readout}. For the edited GRACE model this translates to generating \texttt{output} starting from \texttt{prompt}. 
\item 
\textbf{Concatenate} the sequence of \texttt{output} sentences - here initialized with the first \texttt{prompt} sentence - resulting in a newly-built and model-specific WikiBio entry.
\end{enumerate}

\begin{figure*}[ht]
    \centering
    \includegraphics[width=0.75\textwidth]{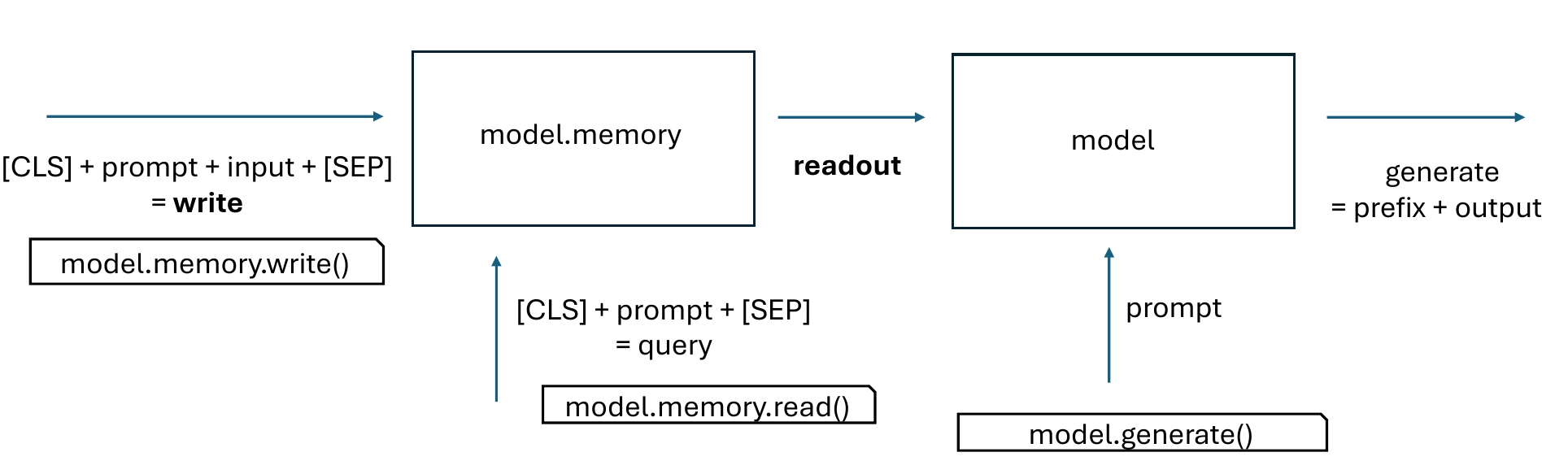}
    \caption{Larimar pipeline for processing (\texttt{prompt}, \texttt{input}) pairs. Here model refers explicitly to Larimar decoder. Larimar encoder is implicitly involved in converting tokens in \texttt{write} and the query \texttt{prompt} (\texttt{prompt} bracketed by \texttt{[CLS]}, \texttt{[SEP]} tokens) into latent vectors.}
    \label{fig:larimar-pipeline}
\end{figure*}

Figure~\ref{fig:larimar-pipeline} illustrates this pipeline for Larimar. 

An actual Wikipedia entry serves as our non-hallucinating baseline. We quantify hallucination in WikiBio entries synthesized by Larimar and GRACE using this baseline as our reference text. Two metrics are employed:
\emph{RougeL} score and \emph{Jaccard} similarity index between reference and synthesized texts. For Jaccard, set operations are computed on tokenized texts: tokenizers respectively from Larimar encoder and GRACE model are used. 

\textbf{Base case: initial explorations.}
By averaging over $i=[238]$ we obtain:

\begin{itemize}
\item
RougeL scores: 
$0.39 \pm 0.14$ for Larimar and  $0.49 \pm 0.18$ for GRACE.
\item
Jaccard similarity scores: 
$0.33 \pm 0.13$ for Larimar and  $0.44 \pm 0.17$ for GRACE.
\end{itemize}

\textbf{Ideal case: $\mathbf{z}_{\texttt{write}} = \mathbf{z}_{\texttt{readout}}$.}
Note that ideally, if latent vector representations for \texttt{readout} and \texttt{write} coincide 
($\mathbf{z}_{\texttt{write}} = \mathbf{z}_{\texttt{readout}}$), the decoder will have the luxury to attend to a (compressed) representation of what was originally written in memory, 
(the (\texttt{prompt}, \texttt{input}) pair). So then, intuitively, when prompted with \texttt{prompt}, the decoder will be effectively constrained towards generating a textual \texttt{output} that will be similar to \texttt{input}, the second element of the pair. But \texttt{input} is the non-hallucinating reference, consequently our metrics will be elevated. This is indeed the case: we get $0.79 \pm 0.13$ for RougeL and $0.72 \pm 0.15$ for Jaccard similarity, that exceed by far GRACE performance. 

\textbf{Partial \texttt{input} case.}
Let's assume that during query we have access to a fraction $f$ of \texttt{input} tokens to augment its \texttt{prompt}. As expected hallucination reduces for increasing $f$. For example, Jaccard similarity on average successively improves from 0.33 (base case; $f=0$) to 0.43 ($f=0.25$), to 0.60 ($f=0.50$), to 0.69 ($f=0.75$), which well exceeed GRACE metrics (0.44) for $f=0.50, 0.75$.

It follows that in Larimar, its memory-computed $\mathbf{z}_{\texttt{readout}}$ vectors provide an additional, unique opportunity for minimizing hallucination in generated output, simply by geometrically aligning them to \texttt{write} encodings. This is not available to SOTA methods for the same task like GRACE, which encapsulate standard, non-memory-augmented models by-design. In what follows we investigate this opportunity.

\textbf{Observations}

\begin{figure}[!ht]
    \centering
    \includegraphics[width=0.65\textwidth]{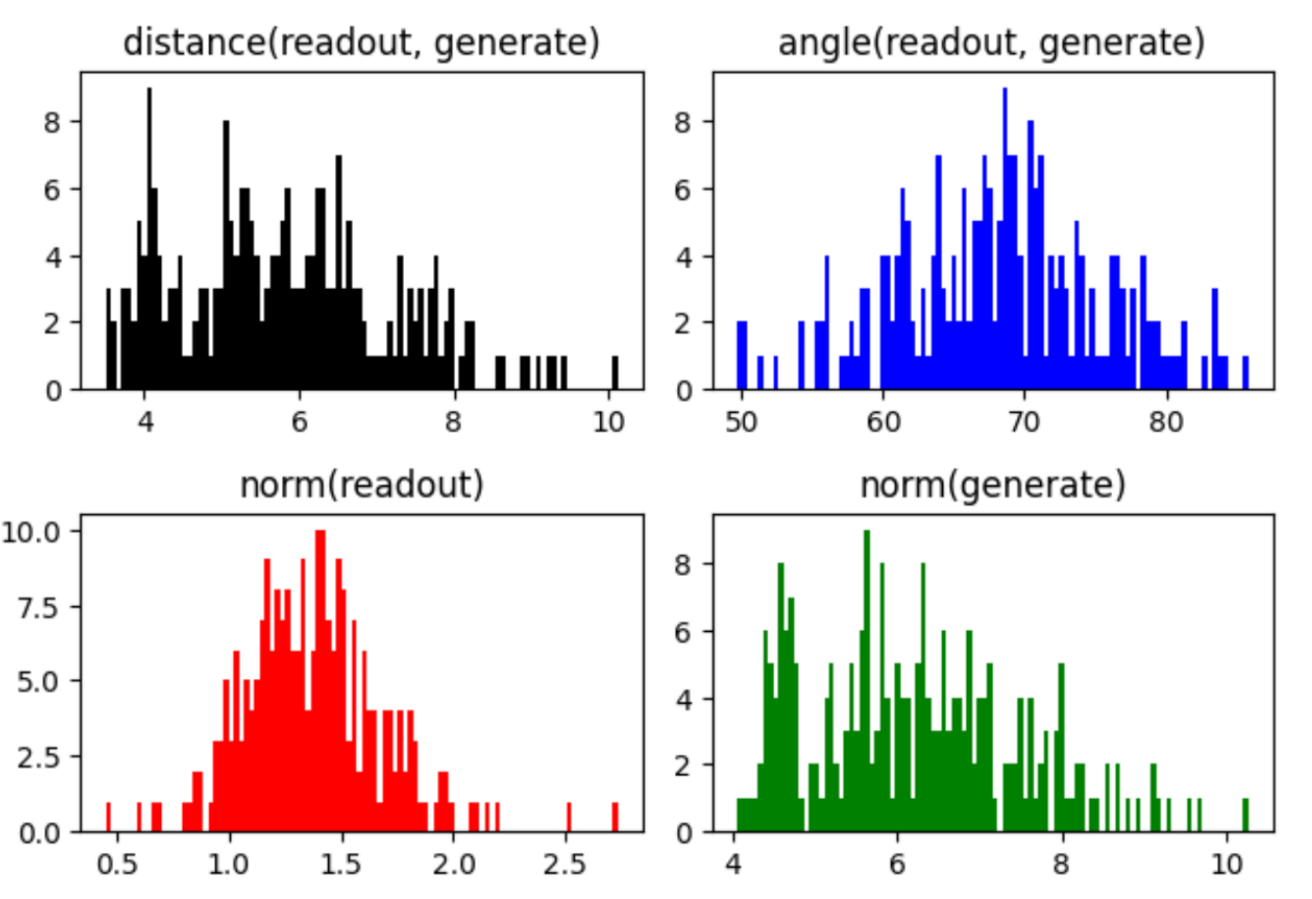}
    \caption{\texttt{readout}, \texttt{generate} pair.}
    \label{fig:readout-generate}
\end{figure}

\begin{figure}[ht]
    \centering
    \includegraphics[width=0.65\textwidth]{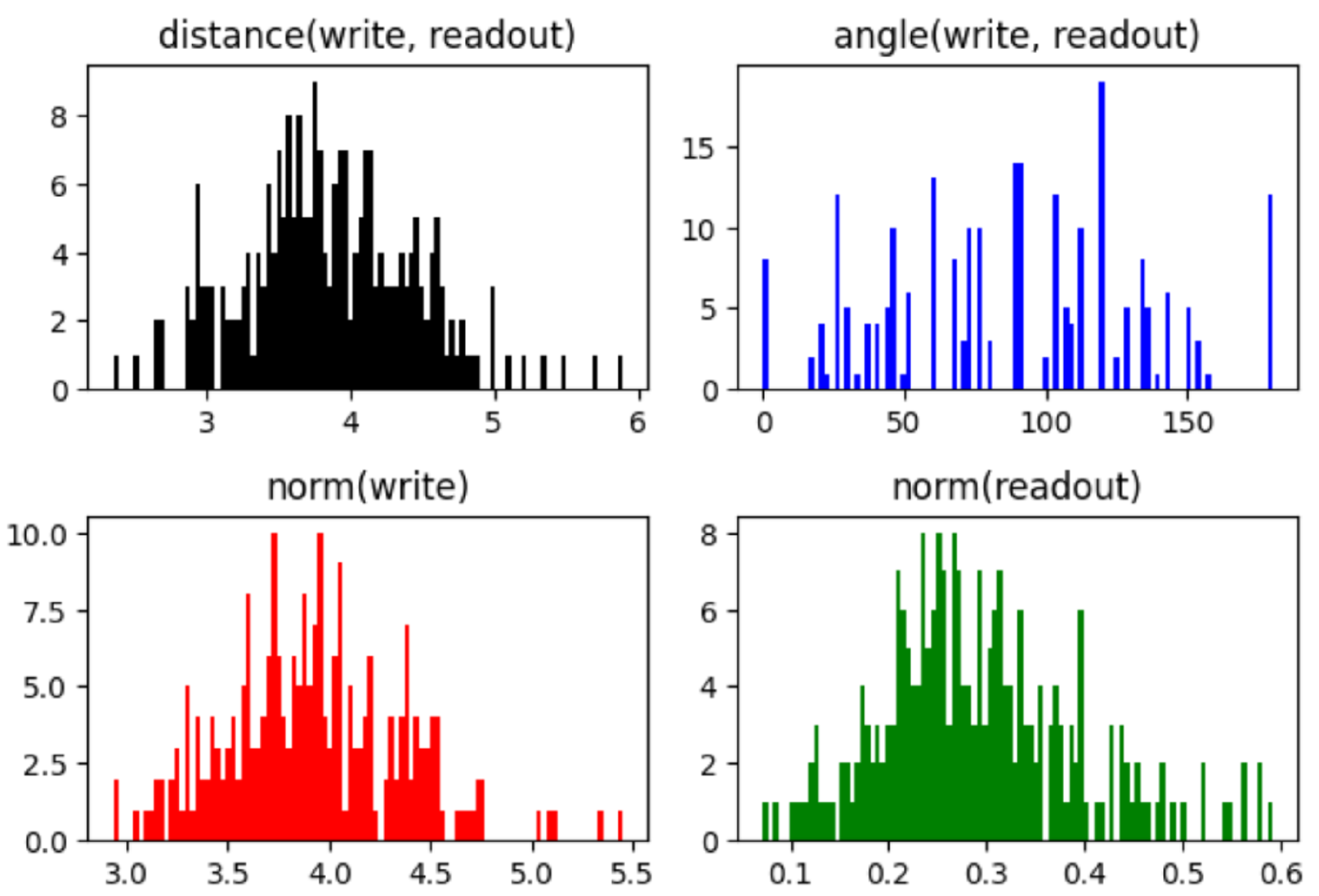}
    \caption{\texttt{write}, \texttt{readout} pair: enforcing the randomly chosen phrase: ``Try to come up with the next wikibio sentence.'' as the query \texttt{prompt}.
}
    \label{fig:write-readout-random}
\end{figure}

\begin{figure}[!ht]
    \centering
    \includegraphics[width=0.65\textwidth]{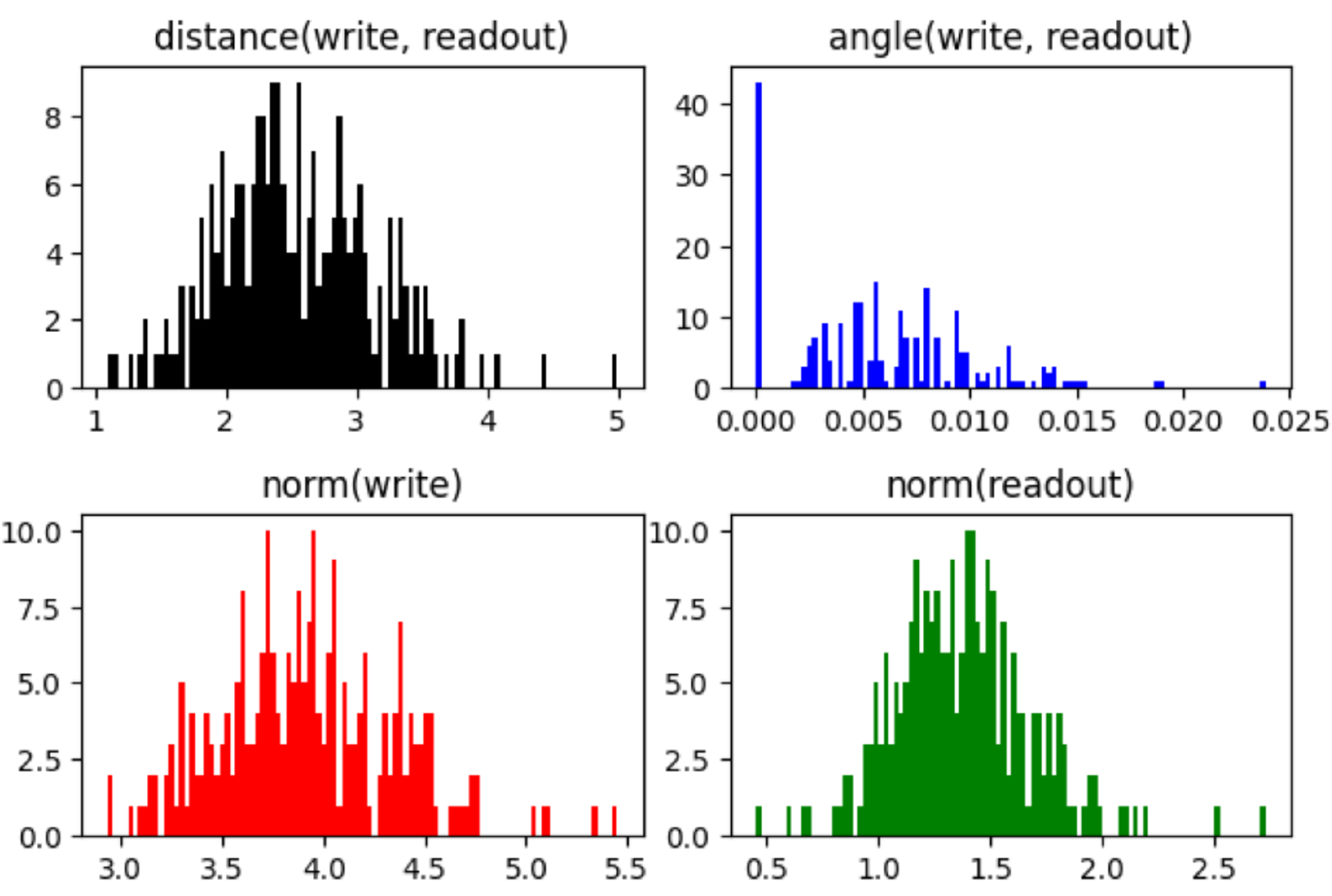}
    \caption{\texttt{write}, \texttt{readout} pair.}
    \label{fig:write-readout}
\end{figure}

Figures~\ref{fig:readout-generate},
\ref{fig:write-readout-random},
\ref{fig:write-readout},
are panels of histograms capturing geometric properties (distance, angle in degrees, $l_2$-norm) of latent vector representations for pairs of texts in various stages of our Larimar pipeline as in Figure~\ref{fig:larimar-pipeline}: x-axis depicts the \emph{average} of the property for each of the 238 biographies, suitaby binned, and y-axis is the count for the bin.

\begin{itemize}
\item
We observe that Larimar decoder arbitrarily distorts both the direction and the magnitude of incoming 
$\mathbf{z}_{\texttt{readout}}$ vectors: $\mathbf{z}_{\texttt{generate}}$ vectors tend to increase in magnitude and deviate over a broad range of acute angles from their decoder inputs. This makes it hard to connect the two vector types (Figure \ref{fig:readout-generate}). Similarly, when we enforce a random prompt in querying the memory - which is a way of muting the constrained generation advantage in Larimar - then $\mathbf{z}_{\texttt{write}}$ and corresponding $\mathbf{z}_{\texttt{readout}}$ vectors significantly deviate from each other (Figure~\ref{fig:write-readout-random}), so we cannot connect them: lengths contract by an approximate factor of $\times 10$ and  their angles distribute wildly over the full $0^{\circ}$-$180^{\circ}$ range.
\item 
There is a clear alignment between $\mathbf{z}_{\texttt{write}}$ and $\mathbf{z}_{\texttt{readout}}$ vectors when standard constrained generation is in effect in Larimar; equivalently when we query its memory with the actual \texttt{prompt} (Figure~\ref{fig:write-readout}). Although there is still a relative decrease in vector length by a factor of $\times 3$ to $\times 4$, 
the two vectors are very well aligned (their angles are tiny fractions of $1^{\circ}$).  
\end{itemize}

This last observation provides a very interesting avenue for constraining generation in Larimar so that hallucination is mitigated: we can scale up the length of $\mathbf{z}_{\texttt{readout}}$ vector by the reported factor (a fixed number $s$ in the range $3$ to $4$ for all samples). Then its distance to $\mathbf{z}_{\texttt{write}}$ can be kept approximately to a minimum and subsequently expect to get hallucination-optimized generations (similar to the ideal case above).

\begin{figure}[!ht]
    \centering
    \includegraphics[width=0.50\textwidth]{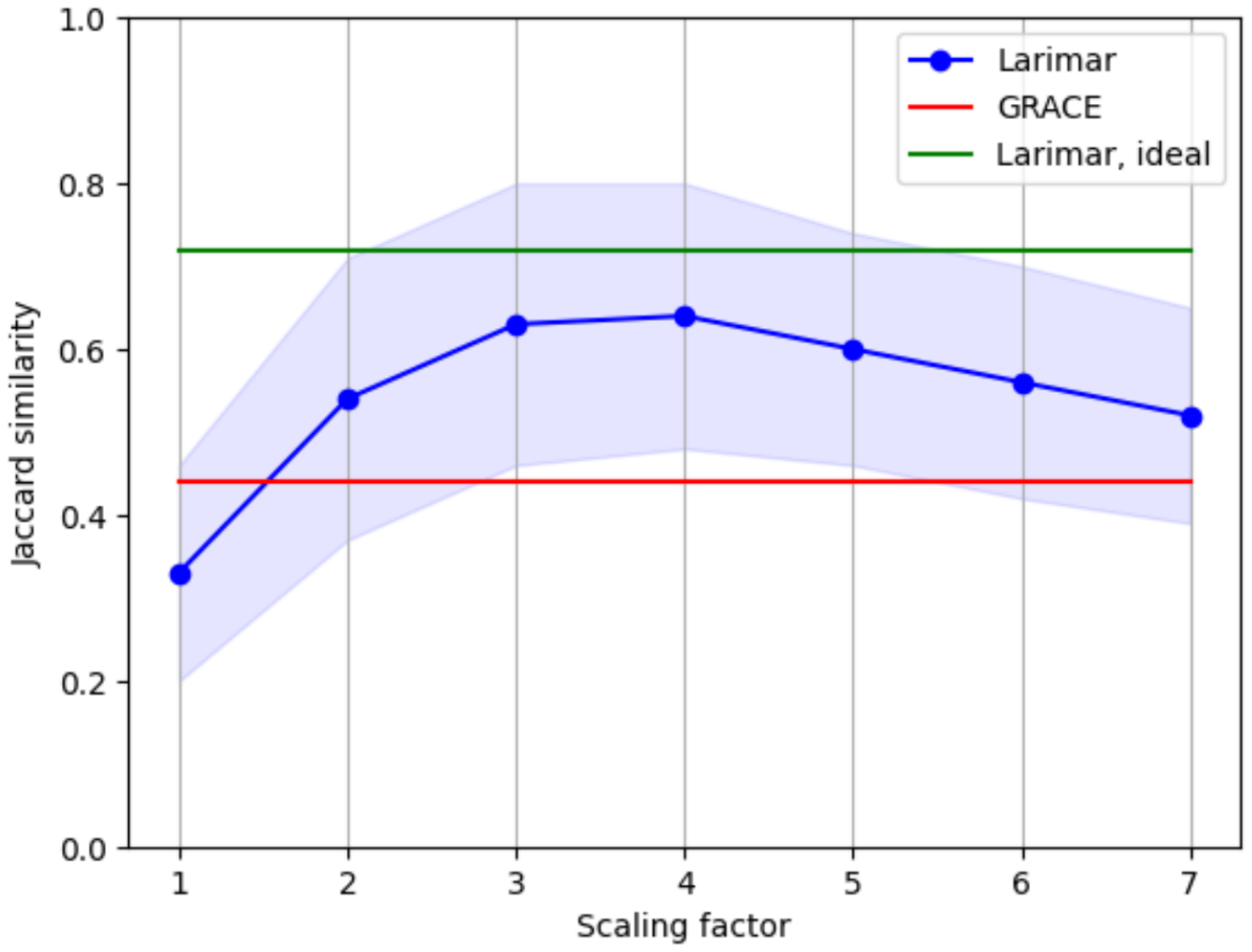}
    \caption{Jaccard similarity for different scaling factors $s$ for $\mathbf{z}_{\texttt{readout}}$ in Larimar. Mean Jaccard similarity scores for the ideal case in Larimar ($\mathbf{z}_{\texttt{write}} = \mathbf{z}_{\texttt{readout}}$)  and for GRACE are also plotted as horizontal lines for comparison.}
    \label{fig:jaccard-similarity}
\end{figure}

Table~\ref{tab:rougel} summarizes mean RougeL scores for the generated output for different scaling factors $s$. We get a maximum value $0.72$ (for $s=4$) which is significantly better than $0.49$ in GRACE ($46.9 \%$ improvement, simply scaling by a single factor all samples). $0.72$ is smaller than RougeL score for the ideal case in Larimar ($0.79$) but this is totally expected since in the ideal case, each sample - implicitly - adapts its scaling factor. We obtain analogous benefits from scaling reflected in Jaccard scores (Figure~\ref{fig:jaccard-similarity}). Both metrics are maximized for $s$ values in the vicinity of $3$ and $4$, exactly as in our observation for the relative contraction of the length of $\mathbf{z}_{\texttt{readout}}$.

\begin{figure}[!ht]
    \centering
    \includegraphics[width=0.55\textwidth]{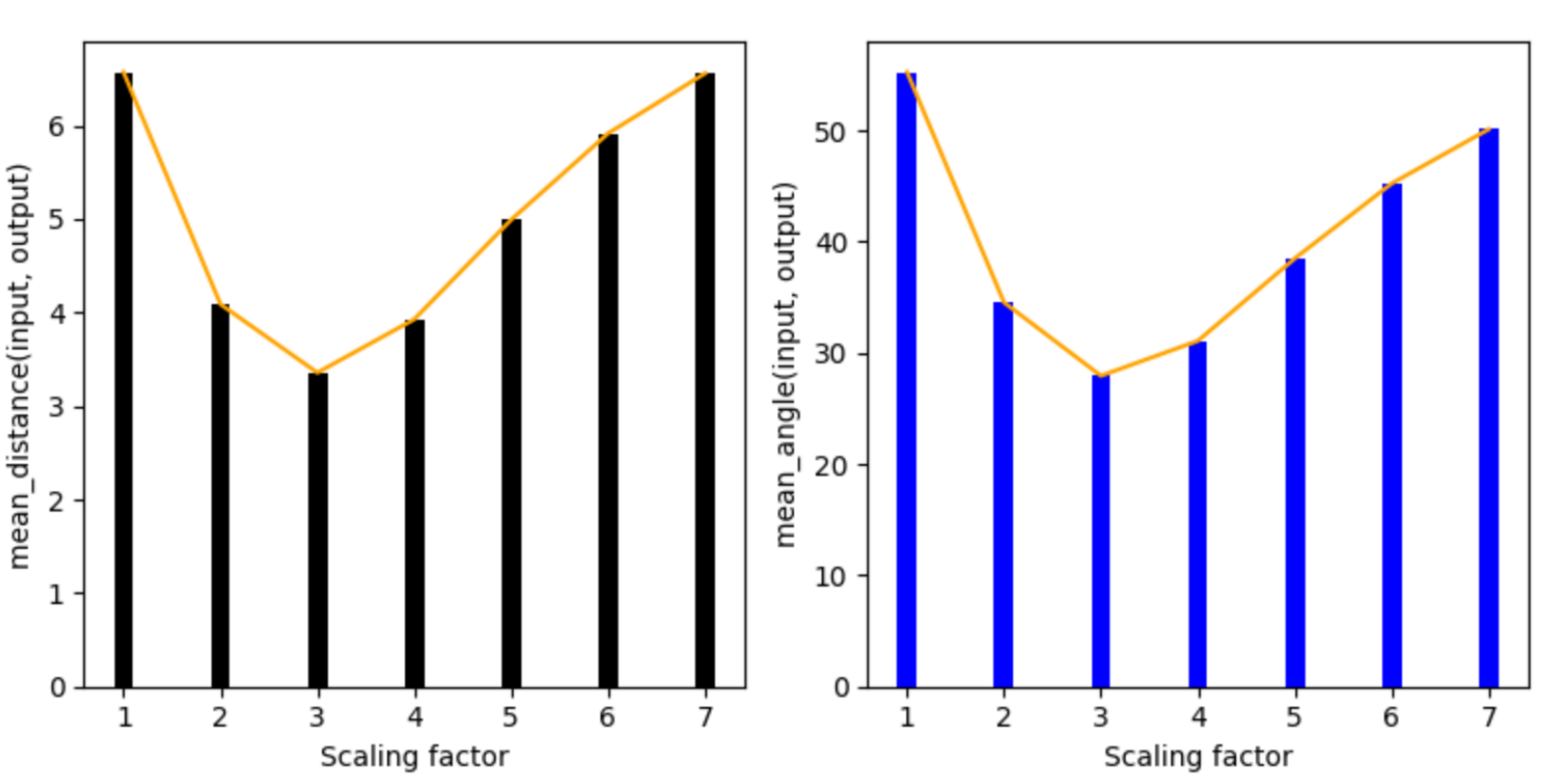}
    \caption{Geometric alignmnent between \texttt{input} and \texttt{output} latent space representations. Mean of vector distances (left) and angles(in degrees; right) for different scaling factors $s$ for $\mathbf{z}_{\texttt{readout}}$.}
    \label{fig:input-output-align}
\end{figure}

Note that enforcing alignment (by scaling) of two vectors, that serve respectively as input and output of a \emph{single} Larimar model module (its memory alone), also optimizes the geometric alignment between two additional vectors, those encoding \texttt{input} and \texttt{output}, with the distingusihing properties: (i) they sit at the two ends of the \emph{full} Larimar model pipeline, (ii) they are not explicitly materialized (for example the encoding for \texttt{write}  - rather than for its enclosed \texttt{input} - is the one explicitly computed). Geometric alignment of the vectors is again optimal in the vicinity of $3$, $4$ for the scaling factor $s$ (Figure~\ref{fig:input-output-align}): on average, distances and angles between them are minimized for this range.

\begin{table}
    \centering
    \begin{tabular}{c l}
        \toprule
        \textbf{Scaling factor ($s$)} & 
        \textbf{RougeL score} \\
        \midrule
$\times 1$ & $0.39 \pm 0.14$\\ 
$\times 2$ & $0.62 \pm 0.17$\\
$\times 3$ & $0.71 \pm 0.16$\\
$\mathbf{\times 4}$ & $\mathbf{0.72 \pm 0.14}$\\
$\times 5$ & $0.69 \pm 0.13$\\
$\times 6$ & $0.64 \pm 0.13$\\
$\times 7$ & $0.59 \pm 1.12$\\
\bottomrule\\
\end{tabular}
\caption{RougeL scores averaged over WikiBio entries for different scaling factors $s$ for the \texttt{readout} vectors, 
$\mathbf{z}_{\texttt{readout}} := s \times \mathbf{z}_{\texttt{readout}}$.}
\label{tab:rougel}
\end{table}

\textbf{Complexity considerations}

GRACE learns codebooks via minimization of a loss function during edits. This is an expensive iterative operation since a number backpropagation steps are necessary.
GRACE model has 1,557,611,200 parameters. It takes 162.5 secs to synthesize a GRACE WikiBio entry. If we do not include the time necessary to reinitialize the model for each pair processing, this can go down to 37.8 secs.

Larimar uses memory writes and reads. These are implemented as  pairs of matrix multiplications (i) for computing the coordinates of distributed memory slots to write to / read from and (ii) for computing the low-rank memory matrix (while writing) or extracting readout representation (while reading). These are lightweight operations.
Larimar encoder has 335,152,128 parameters; Larimar decoder has 810,406,400 parameters. It takes 3.1 secs on average to synthesize a Larimar WikiBio entry.

Model sizes are comparable, with Larimar being slightly smaller (approximately 1.3B versus 1.5B for GRACE). However Larimar is substantially faster in WikiBio synthesis by 1-2 orders of magnitude.

\section{Discussion}
The ability to constrain generation in the decoder by using lightweight memory primitives as in Larimar encoder-memory-decoder architecture, offers an excellent, training-free opportunity for mitigating hallucination. Simple, geometry-inspired operations (here: vector scaling) on selected encodings (here: memory readouts),
although inherently limited to models which, like Larimar, are augmented with explicit memory mechanism,
can be much more effective that training-based approaches (here: GRACE, learning adapter layers parameters by training). 



\newpage
\appendix
\onecolumn

\section{Example of WikiBio generation}

\vspace{0.2cm}
\rule{\textwidth}{2pt}
\textbf{Actual WikiBio entry}\\
Sir John Russell Reynolds, 1st Baronet (22 May 1828 – 29 May 1896) was a British neurologist and physician. Reynolds was born in Romsey, Hampshire, as the son of John Reynolds, an independent minister, and the grandson of Dr. Henry Revell Reynolds. He received general education from his father, and was educated in his profession at University College, London, where he obtained three gold medals in the medical school. In 1851, he graduated M.B. in the University of London, and obtained a scholarship and gold medal in medicine. In 1852, he took the degree of M.D., and began practice in Leeds. He soon after moved to London, and took a house, 38 Grosvenor Street, in which Dr. Marshall Hall had lived. Hall announced to his patients in a printed circular that Reynolds had succeeded him in practice. Such procedure was contrary to a recognised understanding among physicians, and Hall incurred the censure of the College of Physicians.

\rule{\textwidth}{2pt}
\textbf{GPT-3 generated WikiBio entry}\\
John Russell Reynolds (1820–1876) was an English lawyer, judge, and author. He was born in London, the son of a barrister, and was educated at Eton College and Trinity College, Cambridge. He was called to the bar in 1845, and became a Queen's Counsel in 1859. He was appointed a judge of the Court of Common Pleas in 1867, and was knighted in 1871. Reynolds was a prolific author, writing on a wide range of topics. He wrote several books on legal topics, including The Law of Libel and Slander (1863), The Law of Copyright (1865), and The Law of Patents for Inventions (1868). He also wrote on a variety of other topics, including history, biography, and literature. He was a frequent contributor to the Saturday Review, and wrote several books on Shakespeare, including The Mystery of William Shakespeare (1848) and The Authorship of Shakespeare (1875). He also wrote a biography of the poet John Keats (1848).

\rule{\textwidth}{2pt}
\textbf{Larimar}\\
John Russell Reynolds (1820–1876) was an English lawyer, judge, and author. an English lawyer, judge, and author. He was a member of the Royal Society of London and served as Lord Chancellor of Great Britain from 1832 to 1836. born in London, the son of a barrister, and was educated at Eton College and Trinity College, Cambridge. He was a member of the Cambridge Apostles. In 1851, he graduated M. B. in the University of London, and obtained a scholarship and gold medal for medicine. In 1852, he took the degree of M. D., and began practice in Leeds. Slander ( 1863 ), The Law of Copyright ( 1865 ), and The Law of Patents for Inventions ( 1868 ). Hall returned to his Cambridge days in a series of lectures that were published in 1876. several books on Shakespeare, including The Mystery of William Shakespeare ( 1848 ) and The Authorship of Shakespeare ( 1875 ).

\rule{\textwidth}{2pt}
\textbf{Larimar with scaling factor $s=4$}\\
John Russell Reynolds (1820–1876) was an English lawyer, judge, and author. Reynolds was born in Romsey, Hampshire, as John Reynolds, the son of an independent minister, and the grandson of Dr. Henry Revell Reynolds. general education from his father, and was educated at his father's profession in Parliament, where he obtained three gold medals in the London medical school, University became a Queen's Counsel in 1859. in the University of London, and obtained a scholarship and gold medal in medicine. In 1852, he took the degree of M. D., and began practice in Leeds. ). He soon after moved to London, took a room. Marshall Hall, and had lived and Dr Grosvenor engaged in. Hall announced to his patients in a printed circular that Reynolds had succeeded him in practice. Such a procedure was contrary to recognised understanding among physicians, and Hall incurred the censure of the College of Physicians.

\rule{\textwidth}{2pt}
\textbf{GRACE}\\
John Russell Reynolds (1820–1876) was an English lawyer, judge, and author.  He was born in Romsey, Hampshire, and was educated at Repton, Romsey, and Trinity College, Cambridge. He was called to the Bar in 18  He was called to the bar in in in 1989, and became a Queen's Counsel in 1991. He was appointed to the bar in 1992, and became  In 1851, he graduated M.B.  He was a strong and effective advocate for the rights of women and children, and  In 1852, he took the degree of M.D., and began practice in Leeds.  He also wrote extensively on the law of copyright, including The Law of Copyright (1863), The Law of Patents for Inventions (  Hall announced his retirement in a statement to the press. Hall was born in Dublin and  He also wrote extensively on the life of William Shakespeare, including The Life of William Shakespeare (1886), The Mystery


\end{document}